# Improving LLM Classification of Logical Errors by Integrating Error Relationship into Prompts

Yanggyu Lee[1], Suchae Jeong[2], and Jihie Kim[3]

[1] Department of Computer Science and AI, Dongguk University, Seoul, Republic of Korea
[1] yglee730@dgu.ac.kr
[2, 3] College of AI Convergence, Dongguk University, Seoul, Republic of Korea
[2] jeongsuchae9211@gmail.com, [3] jihie.kim@dgu.edu

**Abstract.** LLMs trained in the understanding of programming syntax are now providing effective assistance to developers and are being used in programming education such as in generation of coding problem examples or providing code explanations. A key aspect of programming education is understanding and dealing with error message. However, 'logical errors' in which the program operates against the programmer's intentions do not receive error messages from the compiler. In this study, building on existing research on programming errors, we first define the types of logical errors that can occur in programming in general. Based on the definition, we propose an effective approach for detecting logical errors with LLMs that makes use of relations among error types in the Chain-of-Thought and Tree-of-Thought prompts. The experimental results indicate that when such logical error descriptions in the prompt are used, the average classification performance is about 21% higher than the ones without them. We also conducted an experiment for exploiting the relations among errors in generating a new logical error dataset using LLMs. As there is very limited dataset for logical errors such benchmark dataset can be very useful for various programming related applications. We expect that our work can assist novice programmers in identifying the causes of code errors and correct them more effectively.

## 1 Introduction

In recent developments in Natural Language Processing (NLP), Large Language Models (LLMs) have evolved to understand and infer the meaning of sentences or documents, allowing them to grasp context and understand the relationships between words more accurately. Such approach has significantly boosted the performance of LLMs in various NLP tasks, leading to technical advancements in each NLP domain[1]. In programming, LLMs focus on understanding the workings of source code and performing tasks such as code analysis and autocompletion. Recently, various LLMs, such as codex and codellama[2, 3], have emerged, enhancing the understanding of programming syntax to assist developers in their tasks more efficiently.

While these LLMs serve developers' convenience, there is also the potential for their use in coding education for beginners. For example, students can use LLMs to generate solutions for practice problems not provided by instructors, facilitating more opportunities for code learning[4]. Currently, LLMs are primarily used for code generation[6] and explanation[7]. However, a crucial aspect of programming education is understanding and addressing programming error messages[5]. Understanding the process of interpreting and resolving error messages is necessary to facilitate efficient code learning.

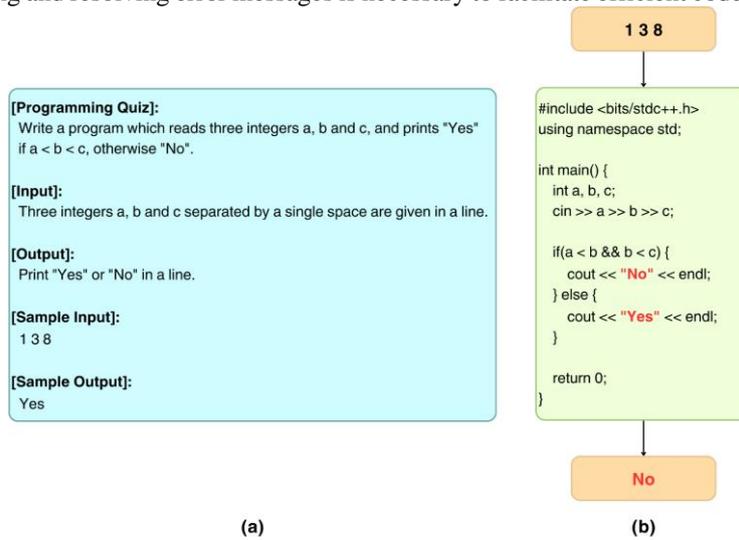

**Figure 1.** The answer is incorrect, but no syntax error occurs when running; (a) is a programming problem that the user must solve; (b) is the incorrect answer code with a Condition Error.

When writing programming code, one particular type of error that requires special attention is logical error. Logic error refers to an error in which a program operates differently from the intention of the program writer and outputs an incorrect result. The significance of logical errors stems from the fact that, unlike compile errors, it is difficult for the novice programmer to self-correct. In the case of syntax errors, the compiler provides error messages, making it feasible to produce corrections. Simply following the code and making fixes enhances the ability to self-debug. However, for logical errors, often times there is no clear error message provided. Therefore, the debugging process involves understanding what the code does, identifying its functionality, and comparing it to the intended process. Novice programmers find this process challenging. **(Figure 1)** illustrates such a scenario where the C++ code written in **(b)** behaves differently from the intended purpose in **(a)**. It is a problem that a program that outputs "Yes" when a specific condition is satisfied by comparing the sizes of the input three integers must be implemented. In (b), the output was wrong by setting the condition differently from the part referred to in (a). That is, the error that occurred in **(Figure 1)** is due to a Condition Error. However, as it is syntactically correct, no error messages are received from the compiler, and the debugging process mentioned above becomes necessary. Detecting such errors and developing an approach for assisting the user holds educational potential **[8]**.

In this paper, building on existing work on program analysis, we first divide the types of errors into ten categories and establish their respective concepts for the purpose of classifying logical errors more clearly. We then identify potential areas of confusion by understanding the relationships between errors and set an ordering in resolving them. Based on the defined concepts and relationships of errors, we propose a new approach for detecting logical errors with LLMs that makes use of relations among error types in the Chain-of-Thought (CoT)[12] and Tree-of-Thought (ToT)[13] prompts. We also undertake the task of generating specific logical error data from the correct code. The contributions of this paper include 1) defining the type of logical error and described the relationship between the types; 2) analyzing relationships between logical error types and reflecting them in LLM prompts, allowing the LLM to clearly distinguish between error types; 3) demonstrating effectiveness of the approach through an experiment where the classification accuracy increases by 21%.

## 2 Related Works

### 2.1 Automated Program Repair

Existing research on errors in source code has predominantly followed the Automated Program Repair (APR) paradigm[9, 16]. One notable example is [9], which introduces a method for fixing a broken program based on compiler-provided error messages. The model generates the corrected lines when given a program with errors along with the corresponding error messages. The work represents the errors and the error messages from the compiler in a graph format, enabling the model to modify multiple lines of code. Moreover, to address the data scarcity issue, the researchers intentionally broke programs for which labels did not exist, obtained error messages from the compiler. However, the approach cannot handle logical errors that do not come with specific error messages, and its focus on code recovery prevents users from understanding the root cause of errors. This paper aims to classify types of logical errors occurring in the code to enable users to understand the reasons behind the errors.

### 2.2 Teaching Programming to Beginners: An Instructor's Perspective on Educational Environment

When teaching coding to beginners, instructors often face the challenge of repeatedly explaining the same errors[14]. During practical sessions, students may spend considerable time fixing simple errors and inadvertently introduce repetitive mistakes. However, due to the large number of students compared to instructors in labs, individual students may need more frequent assistance from instructors. Students are often left to resolve errors independently, relying on error messages for guidance. While compile-time or runtime errors can be addressed to some extent by interpreting compiler-provided error messages, logical errors, which deviate from the intended behavior, often lack explicit error messages, making debugging a particularly upsetting experience. To alleviate these challenges, [17] introduces a prospective solution to enhance the performance of intelligent programming education systems by providing personalized feedback to learners. This study suggests a prospective resolution to mitigate the aforementioned issues. LLMs takes the programming problems as input attempted by students and the submitted code. It then identifies the types of logical errors present and provides this information as output, enabling students to understand the root causes of their errors independently. Understanding error messages is crucial for effective programming learning[5], and use of LLM based approaches could benefit programming novices.

## 3    Definition of logical error types

In the past, there has been research on a classification system for common logical errors made by novice programmers[10]. The 11 error classifications explained in [10] list examples for each error type. However, a clear definition of the error type is not fully provided. In our research, we first present a definition of individual errors (**Table 1**) for classification of logical errors. In order to cover logical errors in diverse programming languages, components of specific grammar such as class programming were excluded. The classification system we reconstructed can be applied to languages such as C, Java, and Python.

**Table 1.** Group types commonly occur while writing a code; Column 1 and 2 are list error types and names; Column 3 provides a description and examples of the error type

| Type | Category | Description and occurrence examples |
|------|----------|-------------------------------------|
| (A) | Input | Errors arising from the inability to properly store input values properly.<br>1.    When not all given input values are received.<br>2.    When the data type of the variable the input value is incorrect. |
| (B) | Output | Errors arising from non-compliance with the required output format.<br>1.    When the output format of the value in the variable is incorrect.<br>2.    When an incorrect string literal is output. |
| (C) | Variable | Errors arising from incorrect use of variables.<br>1.    When the value stored in the variable is incorrect.<br>2.    When the data type of the variable is incorrectly specified. |
| (D) | Computation | Errors caused by incorrect calculations.<br>1.    When calculating using incorrect values.<br>2.    When calculating using incorrect operations. |
| (E) | Condition | Errors caused by incorrect use of conditional statements.<br>1.    When the conditional operation in the declaration part of the conditional statement is incorrect.<br>2.    When the condition in the declaration part of the conditional statement is insufficient. |
| (F) | Branching | Errors caused by incorrect branching of the program.<br>1.    When the break in the loop is written incorrectly.<br>2.    When a conditional statement that should be written as if-else is written as if-if. |
| (G) | Loop | Errors caused by incorrect use of loops.<br>1.    When the condition in the declaration part of the loop is incorrect.<br>2.    When the variable used in the declaration part of the loop is incorrect. |
| (H) | Array/String | Errors caused by incorrect arrays or strings.<br>1.    When arrays or strings are initialized incorrectly. |

|     |           | 2. When referencing an incorrect index when using an array or string. |
| --- | --------- | --- |
| (I) | Function  | Errors caused by incorrect user-defined functions.<br>1. When the parameters or return values of user-defined functions are incorrectly defined.<br>2. When the arguments are incorrect when calling user-defined functions. |
| (J) | Conceptual | Errors caused by incorrect concepts for problem-solving.<br>1. When solving a different problem than the one presented.<br>2. When the necessary loops or conditional statements are not written to solve the presented problem. |

The '(11) Miscellaneous' category of errors defined in **[10]** encompasses issues arising from incorrectly placed semicolons or minor typos. These errors resemble compilation errors, but they can also be logical errors. When semicolons are misused, the compiler can interpret it as a long sentence. Also, typos in variable names can mislead the compiler. In such cases, it is difficult to set clear criteria and can be easily confused with other types of errors, so the category has been excluded from consideration.

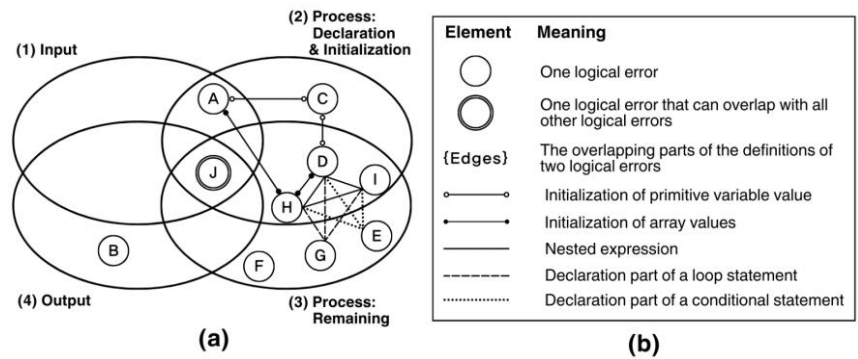

**Figure 2.** Relationship diagram for types of logical errors; (a) divides error types into four areas and displays the relationships between the types in a graph form; (b) refers to the element in the graph of (a).

Next, we created a view **(Figure 2)** to intuitively present the occurrence positions in the code and the relationships among the ten logical errors. Here, each type of logical error is represented as a node with the Type ID defined in **(Table 1)** written on it.

We classified the structure of programs into 'Input', 'Process', and 'Output', according to the IPO pattern that many simple programs follow**[15]**. Subsequently, we distinguished the part used for declaration and initialization in Process and classified logical error types into four major groups: (1) Input, (2) Process: Declaration & Initialization, (3) Process: Remaining, and (4) Output. And we separated the relationships between errors into casual and coincidence relationships.

Casual relationships are represented by the sequence of the structure of programs 'Input', 'Process', and 'Output'. If an error occurs in one step, it may lead to subsequent errors in the following steps. For instance, if a program receives incorrect input in (1)

Input, it can trigger errors in (2) Process: Declaration & Initialization, (3) Process: Remaining, and (4) Output. These casual relationships can obscure the logical error that needs to be classified, as one error can lead to another.

'Coincidence' relationships refer to when the criteria for classifying an error occurrence into a certain error type are ambiguous because the conceptual meaning of two errors overlap. For example, an incorrect operation written in the declaration part of a conditional statement falls under both the Computation error and Condition error categories. If the existence of the conditional statement itself is incorrect, it also falls under the Conceptual error category. These coincidence relationships can make the criteria for error classification unclear.

In (**Figure 2**), you can check the casual relationships according to the (1), (2), (3), and (4) groups where the nodes are located and the coincidence relationships through the edges between each node. You can also see the typical cases where coincidence relationships occur through the meanings of each edge. In this case, (J) Conceptual error can occur, overlapping with all other errors conceptually, so we depicted it as a node with two external lines.

Distinguishing and defining error types based on these casual and coincidence relationships enables us to identify potential confusion when multiple errors could be blended. This confusion can create ambiguity, making it difficult to pinpoint a single logical error in the code or leading to incorrect error classification. Providing clear criteria for such ambiguities enhances error classification performance using LLMs. Particularly, as seen in (Figure 2), there are complex coincidence relationships between nodes (D), (E), (G), (H), and (I). Therefore, it's essential to distinguish the criteria for this area clearly.

While we can specify which error will occur in each relationship to define the relationships between errors, setting and remembering this for all relationships is challenging. Therefore, we assign occurrence priorities to each error to be more applicable in general cases. To ensure convenience, we initially set the highest ordering for (J), which can overlap with all errors and concepts. Then, based on the casual relationships of the errors, we establish the order of arrangement as (J) > (A), (C) > (D), (E), (F), (G), (H), (I) > (B). Subsequently, we adjust the ordering of errors with clearly defined occurrence points among those having coincidence relationships, resulting in the order of arrangement (J) > (A) > (C) > (H) > (I) > (E) = (G) > (D) > (F) > (B). This approach allows us to identify the highest-ranked error when classifying a single logical error to which multiple concepts apply, thus helping to avoid ambiguous situations. Therefore, LLMs can classify logical errors on a more precise basis. Through this, we can inform novice programmers of the incorrect logic in the code more accurately.

## 4    Classification & Augmentation Using LLMs

The structure of the data used for classification is as follows. The labeling format is binary, with a value of 1 indicating the presence of the corresponding error type in the code and 0 indicating its absence. The distribution of the collected data is presented in (**Table 2**). The data can be access on GitHub.[1]

---

[1] https://github.com/SChaeck/llm-logical-error-detection

**Table 2.** Distribution of logical error types collected for classification. The first row is the logical error type, and the second row is the number of data per logical error type.

| (A) | (B) | (C) | (D) | (E) | (F) | (G) | (H) | (I) | (J) |
|-----|-----|-----|-----|-----|-----|-----|-----|-----|-----|
| 10  | 10  | 5   | 9   | 12  | 8   | 10  | 8   | 8   | 6   |

Additionally, the data used for augmentation was only the code data that had been 'Accepted' as the correct answer for problems in the Introduction to Programming 1 (ITP 1) course.

### 4.1 Logical Error Classification Prompt

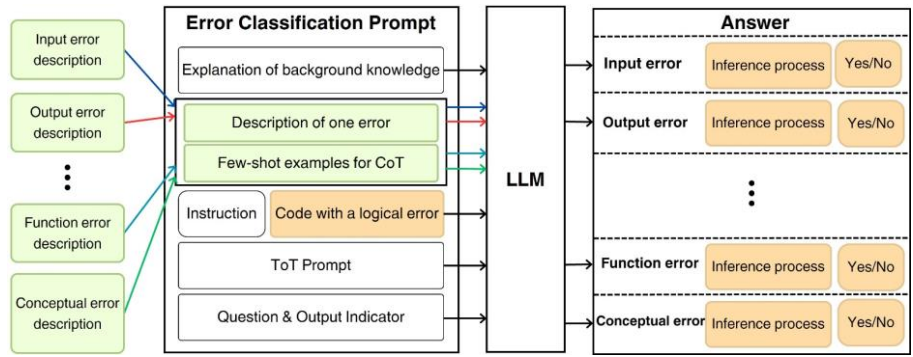

**Figure 3.** This is the overall structure and process for classifying logical error types from incorrect answer codes.

We constructed a pipeline to observe the logical error classification performance of LLMs. **(Figure 3)** illustrates the process of the LLM classifying logical errors based on the provided information.

The Error Classification Prompt is composed of 'Explanation of background knowledge', 'Description of one error', 'Few-shot examples for CoT', 'Instruction', 'Code with a logical error', 'ToT Prompt' and 'Question & Output Indicator'.

The 'Explanation of Background Knowledge' section explains the model of what a logical error is and lists the names of ten errors that will be used as classification criteria. This gives the LLM preliminary information on what task it needs to perform. 'Description of one error' uses ten descriptions written based on the concept of errors defined in **(Table 1)** to describe each type of logical error. As only one error content exists in a single description, it's impossible to inform the LLM of the concept of all errors. Therefore, by writing about the restrictions on the occurrence of the described error, we have expressed its relationship with other errors. Through this, we provided cases where the concept of the error applies but should not occur. 'Few-shot examples for CoT[12]' provides instances for each error type. Each example consists of three average shots for each type of error. This helps the LLM distinguish the ten error types and understand where to focus. In 'Instruction', it asks the LLM if the previously explained error types exist in the code. 'Code with a logical error' is formatted as in **(Figure 1)**, consisting of the problem the programmer wants to solve and the incorrect code for that problem. Finally, ToT prompts[13] are written to facilitate three experts sharing their thoughts step by step and providing feedback on each other's ideas, followed by 'Question &

Output Indicator.' This enhances the accuracy of answers for complex errors that require inference.

Therefore, when one incorrect code is input, it generates ten prompts that verify different errors for this code, prompting the LLM with questions. Only the description changes during this process, while the rest remains unchanged. The LLM explains the inference process for each of the ten prompts and responds with Yes/No, indicating whether the specific error is present. This allows for identifying which logical errors exist in a given code.

## 4.2   Logical Error Augmentation Prompt

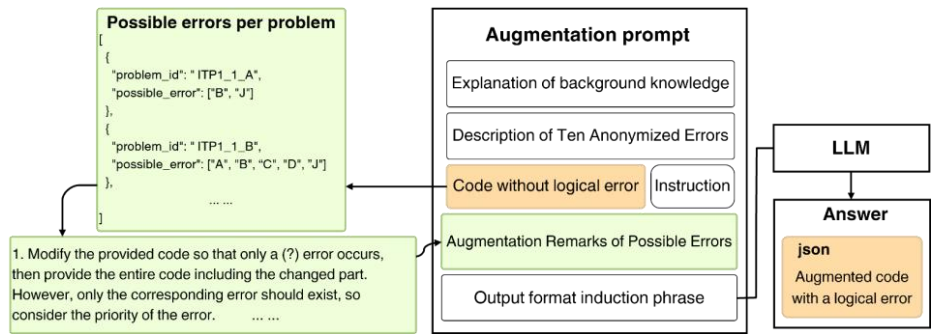

**Figure 4.** Process and configuration of generating a code in which a logical error occurred from code with an 'Accepted' judgment.

We proposed a methodology for constructing a logical error dataset by creating augmentation prompts to induce specific logical errors from correct answer data. The overall process is illustrated in **(Figure 4)**.

The augmentation prompt consists of an 'Explanation of background knowledge', 'Description of Ten Anonymized Errors', 'Code without logical error', 'Instruction', 'Augmentation Remarks of Possible Errors', and 'Output' format induction phrase'. In the 'Explanation of background knowledge' of the augmentation prompt, unlike the classification prompt, we explained what logical errors are. This was done as each error's enumeration is described with explanations in the subsequent 'Description of Ten Anonymized Errors'. The 'Description of Ten Anonymized Errors' provides detailed descriptions and scenarios for each error, using alphabet anonymization instead of direct names for all ten errors. This prevents augmentation based on the content of each error stored in the parametric memory, enabling a focus on the error classification concepts in the research. For instance, to avoid focusing solely on the term 'Input error' rather than the 'explanation for the Input error', the error type is provided to the LLM in the form of (A). Also, by providing the ordering of anonymized errors and explanations about the order, we ensured that the LLM considers the relationship between errors, preventing augmentation with incorrect errors. 'Code without logical error' comprises problem descriptions and correct code, similar to **(Figure 1, (a))**. The instruction assigns the task of augmenting while including common mistakes novice programmers make. 'Augmentation Remarks of Possible Errors' only requests errors that can be augmented. Given the varying difficulty levels and unused syntax in the collected dataset problems, potential errors are compiled, and 'Augmentation Remarks of Possible Errors' are created based on the 'Code without logical error' problems. In 'Output format induction phrase,' the LLM is instructed to provide a JSON-formatted response. Subsequently, the LLM returns the code containing logical errors in JSON format based on the input prompts. This approach allows the construction of a dataset containing logically erroneous code examples. Augmentation results can be checked in (**Figure 5**).

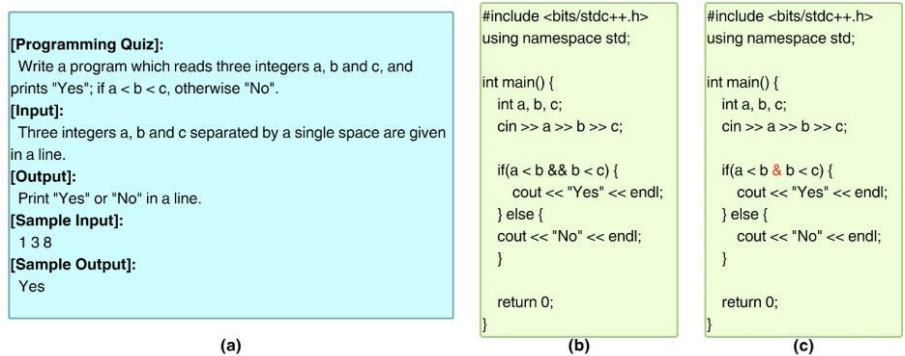

**Figure 5.** Generation of code in which a specific logical error occurred from code with an 'Accepted' judgment; (a) is the problem we want to solve; (b) is code for 'Accepted' judgement; (c) is code where a logical error was generated.

### 4.3 Experimental Results

**Method of collecting experimental data.** The labeled logical error dataset was relatively limited, requiring us to collect and label additional data manually. Then, from the data collected, we only utilized those labeled with the agreement of two annotators. The labeled data was used to evaluate the accuracy of Large Language Models (LLMs) in classifying logical error types and for augmentation tasks. The raw data was collected from the programming problem-solving platform AOJ (Aizu Online Judge) [11]. Specifically, we focused on problems from the ITP 1 course, which is part of AOJ's introductory programming courses. When users submit their solutions to programming problems and the system evaluates them, the submission information, including submission ID, submitter, submission status, and submission code, is stored on the server. This data is publicly accessible in API format.

**Results.** Looking at (**Table 3**), we can see that the classification accuracy improves when a description is provided for errors, compared to when the description is not provided. When utilizing GPT-3.5-turbo, the classification accuracy for the prompts labeled 'Description not provided' was 35%. However, for the prompts categorized under 'Description provided,' the accuracy improved to 56%, indicating a 21% increase in classification performance. Additionally, we observe that the classification accuracy improves as the model's parameter increases. The performance improvement of GPT-4 compared to GPT-3.5-turbo is attributed to the difference in inferential capabilities based on the model's size. It is presumed that GPT-4 better understands the criteria for errors and can deeply contemplate changes caused by code execution. The False Positive Rate (FPR) is the ratio of instances where 'Yes' is output for incorrect errors when classifying the ten prompts generated from a single code. As errors are more effectively distinguished and classified, the FPR decreases. Therefore, the clearer the limitation of the error written in the description, the more distinguishable the relationship between errors will be, and the FPR will decrease. Additionally, this reduction can be seen in association with the improved inferential capabilities of the model.

**Table 3.** Logical error classification results

| | GPT-3.5-Turbo | | | GPT-4 | |
| | Description not provided | Description provided | | Description provided | |
| **Error Type** | **Acc** | **Acc** | **FPR** | **Acc** | **FPR** |
|---|---|---|---|---|---|
| Input | 20% (2/10)) | 50% (5/10) | 0.06 | 100% (10/10) | 0.1 |
| Output | 20% (2/10) | 80% (8/10) | 0.1 | 100% (10/10) | 0.08 |
| Variable | 40% (2/5) | 60% (3/5) | 0.22 | 100% (5/5) | 0.12 |
| Computation | 67% (6/9) | 44% (4/9) | 0.12 | 55% (5/9) | 0.12 |
| Condition | 50% (6/12) | 67% (8/12) | 0.25 | 92% (11/12) | 0.125 |
| Branching | 38% (3/8) | 50% (4/8) | 0.188 | 63% (5/8) | 0.188 |
| Loop | 50% (5/10) | 50% (5/10) | 0.18 | 100% (10/10) | 0.1 |
| Array/String | 25% (2/8) | 50% (4/8) | 0.138 | 88% (7/8) | 0.275 |
| Function | 25% (2/8) | 63% (5/8) | 0.038 | 88% (7/8) | 0.075 |
| Conceptual | 0% (0/6) | 33% (2/6) | 0.167 | 67% (4/6) | 0.15 |
| **AVG** | **35% (30/86)** | **56% (48/86)** | **0.145** | **86% (74/86)** | **0.13** |

Analyzing the results of GPT-4, we find that despite having relatively high inferential capabilities, it struggles to easily detect errors in areas that conceptually need more clarity, such as 'Computation error,' 'Branching error,' and 'Conception error.' 'Computation error' often occurs in places with incorrect operations, 'Branching error' in locations with incorrect branching, and 'Conception error' where the logic is flawed. Such abstract descriptions can confuse the LLM, making it challenging to understand which part of the code requires attention. On the other hand, errors clearly classified as 'Input error,' 'Output error,' 'Variable error,' 'Condition error,' and 'Loop error' have well-defined locations, namely in the input statement, output statement, variable creation, condition statement declaration, and loop declaration, respectively. This indicates that the model can easily classify errors occurring only in specific parts of the code but struggles with errors present in ambiguous areas.

In (**Table 4**), the augmented results are shown in (**Section 4.2**). A total of 111 code samples were augmented using gpt-3.5-turbo. 'Right Augmentation' refers to cases where augmentation for a specific type of logical error was done correctly. The rest fall under 'Wrong Augmentation,' which is further divided into 'Other type of logical errors' and 'Not a logical error.' 'Other type of logical errors' pertains to cases where the code was generated with logical errors not requested. In contrast 'Not a logical error' denotes cases where a compilation error occurred, or the code provided the correct answer.

Overall, a higher augmentation success rate was observed for errors that are relatively easier to transform, such as (A) Input, (B) Output, and (J) Conceptual error. In contrast, for errors with lower ordering, like (D) Computation and (F) Branching error, more codes augmented into different errors were observed than the number of correctly generated codes. Interestingly, (E) Condition error showed a very low success rate, with

5 out of 8 'Not a logical error' cases resulting in compilation errors and 2 cases producing the same code as the original. This suggests that appropriate prompt engineering could improve the success rate.

**Table 4.** The result of augmentation with a specific type of logical error

| Error Type | Number of Augmented Codes | Right Augmentation | Wrong Augmentation | |
|---|---|---|---|---|
| | | | Other type of logical errors | Not a logical error |
| Input | 10 | 9 | 1 | 0 |
| Output | 13 | 11 | 1 | 1 |
| Variable | 10 | 4 | 2 | 4 |
| Computation | 10 | 1 | 6 | 3 |
| Condition | 10 | 1 | 1 | 8 |
| Branching | 12 | 3 | 4 | 5 |
| Loop | 12 | 5 | 2 | 5 |
| Array/String | 11 | 4 | 2 | 5 |
| Function | 10 | 4 | 2 | 4 |
| Conceptual | 13 | 7 | 3 | 3 |
| **Total** | **111** | **49** | **24** | **38** |

An additional 73 code sets consisting of 49 "Right Augmentation" and 24 "other types of logical errors" can be utilized to evaluate the performance of the error classification model.

## 5 Conclusion

In this paper, we defined ten concepts of logical error types based on prior research and set an order to eliminate potential confusion from understanding the relationship between errors. Then, based on these clearly defined errors, we instructed GPT to classify these errors using CoT and ToT techniques, and evaluated its performance using a manually created dataset. Additionally, we proposed a methodology for creating a benchmark dataset by augmenting the correct code to generate logical errors using each error's definition. We also observed that as the model parameters increased, the inference performance improved, and a clearer redefinition of error types led to a higher classification performance.

**Limitations.** The limitation of this research lies in the insufficient consideration given to the unique characteristics of each programming language, as the newly defined category of logical errors has been applied across various programming languages. This makes it challenging to detect logical errors resulting from the misuse of language-specific syntax, such as class-based programming or pointers. Therefore, a more specialized approach that considers these language-specific characteristics is needed.

**Acknowledgments.** This research was supported by the MSIT(Ministry of Science and ICT), Korea, under the ITRC(Information Technology Research Center) support program(IITP-2024-2020-0-01789), and the Artificial Intelligence Convergence Innovation Human Resources Development (IITP-2024-RS-2023-00254592) supervised by